# *INCIDENCE CALCULUS:*
# *A MECHANISM FOR PROBABILISTIC REASONING*

Alan Bundy

Department of Artificial Intelligence,
University of Edinburgh,
Hope Park Square,
Edinburgh, EH8 9NW, Scotland.

**Abstract**
Mechanisms for the automation of uncertainty are required for expert systems. Sometimes these mechanisms need to obey the properties of probabilistic reasoning. A purely numeric mechanism, like those proposed so far, cannot provide a probabilistic logic with truth functional connectives. We propose an alternative mechanism, **Incidence Calculus**, which is based on a representation of uncertainty using sets of points, which might represent situations, models or possible worlds. Incidence Calculus does provide a probabilistic logic with truth functional connectives.

**Acknowledgements**
I am grateful for comments and advice on this paper from Roberto Desimone, Richard O'Keefe, David Spiegelhalter, Bernard Silver, Steve Owen, M.J.R. Healy, D.M. Titterington, Allen White, Chris Robertson, Peter Fisk and Alberto Petterossi.

An earlier and longer draft of this paper, [Bundy 84], appeared in the proceedings of the International Conference on Fifth Generation Computer Systems, Tokyo, 1984. A revised, longer version will appear in the Journal of Automated Reasoning.

This research was supported by SERC/Alvey grant, number GR/C/20826.

## 1. Introduction

Several mechanisms have been suggested for the automation of reasoning with uncertainty, e.g. Fuzzy Logic, [Zadeh 81], Shafer-Dempster theory, [Lowrance & Garvey 82], and the mechanisms proposed in MYCIN, [Shortliffe 76] and PROSPECTOR, [Duda et at 78]. Most of these mechanisms involve assigning numbers to axioms (e.g the facts and rules of an expert system), and assigning arithmetic functions to the rules of inference, so that new numbers can be calculated for the theorems that are derived from the axioms (e.g. the diagnoses of an expert system). We will call such mechanisms, **purely numeric**.

We will see that it is a desirable property of any uncertainty mechanism that the connectives should be **truth functional** with respect to the uncertainty measures. That is, it should be possible to calculate the uncertainty measure of a complex formula solely from the uncertainty measures of its subformulae. Note that the Propositional Logic connectives are truth functional with respect to the truth values true and false.

In some applications it is important to be able to assign *meaning* to the numbers so obtained, rather than use them merely to rank order some options. For instance, in medical diagnosis a user sometimes needs to be able to distinguish the situations where a diagnosis is very probably correct from the situation where it is just the best of an improbable batch. In the first case a surgeon might be prepared to perform a dangerous operation, in the second s/he might want to call for more tests and a re-diagnosis. In these situations we would like the numbers to represent probabilities.

Unfortunately, the logical connectives *cannot* be truth functional with respect to probabilities, or any other purely numeric uncertainty measure (see [Bundy 84]). We will propose a mechanism, called **Incidence Calculus**, in which the uncertainty measures are sets, for which the connectives *are* truth functional, and from which probabilities can easily be calculated.

## 2. A Set Theoretic Mechanism

Incidence Calculus is based on the set-theoretic roots of probability theory. We will associate uncertainty values with **sentences**, i.e. formulae with no free variables, in some logic, e.g. Predicate Logic. The probability of a sentence is based on a **sample space** of **points**, [Feller 68]. Each point can be regarded as a situation, Tarskian interpretation or possible world in which a sentence will be either true or false. The sample space is intended to be an exhaustive and disjoint set of points. It will be denoted **w**.

Non-trivial theories often have an infinite number of possible interpretations. For computational reasons we will usually require the sample space to be finite. Thus each point must sometimes stand for a, possibly infinite, equivalence class of interpretations.

Let $i(A)$ be the subset of w, containing all those points in which sentence A is true. We will call $i(A)$, the **incidence*** of A. In [Feller 68] what we call an incidence is called an event, however, the term event is also used, ambiguously, to refer to sentences. The distinction between incidences and sentences is crucial in the discussion below, so we will not use the ambiguous term 'event'.

The dependence or independence of two sentences is coded in the amount of intersection between their incidences. The amount of intersection of two independent sentences is no more or less than you would expect from a random assignment of the elements of their incidences.

For the rest of this paper we will assume the underlying logic to be Predicate Logic. The following axioms of Incidence Calculus associate a set theoretic function with each connective, propositional constant and quantifier of Predicate (Propositional) Logic so that the incidence of a complex sentence can be calculated from the incidences of its subsentences. We call the resulting system **Predicate (Propositional) Incidence Logic**.

$$i(t) = w \qquad \text{(i)}$$

---

*incidence: degree, extent or frequency of occurrence; amount. - Collins English Dictionary.

$$i(t) = w \qquad \text{(i)}$$
where $t$ is universally true sentence

$$i(f) = \{\} \qquad \text{(ii)}$$
where $f$ is universally false sentence

$$i(\sim A) = w \setminus i(A) \qquad \text{(iii)}$$

$$i(A \ \& \ B) = i(A) \cap i(B) \qquad \text{(iv)}$$

$$i(A \lor B) = i(A) \cup i(B) \qquad \text{(v)}$$

$$i(\forall X \ A(X)) \subseteq i(A(s)) \subseteq i(\exists X \ A(X)) \qquad \text{(vi)}$$
where $s$ is a term containing no variables not bound in $A(X)$

Note that there is no independence condition on equations (i) to (v), and hence that all the connectives are truth functional with respect to incidences. The axioms for the universal and existential quantifiers only set upper and lower bounds, respectively, on their incidences, so the quantifiers are not truth functional with respect to incidences. This is not surprising since they are not truth functional with respect to truth values either.

If $J$ is a point, let $\varrho(J)$ be the probability of $J$ occurring. If $I$ is a set of points, let wp(I) be the sum of the probabilities of the points in $I$, i.e.

$$wp(I) = \sum_{J \in I} \varrho(J)$$

wp(I) is called the **weighted probability** of I. For computational reasons we will usually use finite I, but the theory does not require I to be finite or even discrete.

Since the sample space is meant to be an exhaustive and disjoint set of points, we will require that:

   wp(w) = 1

If A is a sentence, let p(A) be the **probability** of A occurring. We define

   p(A) = wp(i(A))

If A and B are sentences, let p(A|B) be the **conditional probability** of A given B. We define:

   p(A|B) = wp(i(A)∩i(B))/wp(i(B))

From these definitions it is easy to derive the equations for probability.

   p(t) = 1

   p(f) = 0

   p(~A) = 1 - p(A)

179

p(A & B) = p(A).p(B)
          provided A and B are independent

p(A v B) = p(A) + p(B) - p(A & B)

p(A & B) = p(B).p(A|B)

## 3. The Representation of Incidences

One of the advantages of a purely numeric mechanism for uncertainty is that computers are particularly efficient at representing and manipulating numbers. They are not so efficient at representing and manipulating sets.

However, Incidence Calculus can be implemented reasonably efficiently by representing the incidences of sentences as bit strings and manipulating them with logical operations. Each incidence can be represented by a bit string of a fixed length, say 100 bits, each bit corresponding to an element of w. The longer the string, the greater the accuracy, but the greater the cost in terms of space and time. Each bit in a string is 1 or 0 according to whether the element it corresponds to is or is not in the incidence being represented.

The incidence of A & B can then be calculated by taking the logical *and* of the incidences of A and B; the incidence of A v B can be calculated by taking the logical *or* of the incidences of A and B; and the incidence of ~A can be calculated by taking the logical *not* of the incidence of A.

Incidences can be a more efficient method of storing probabilistic information than probabilities. Consider the Propositional Incidence Calculus situation and suppose that we want to assign the probabilities of a set of sentences. Let this set contain n different propositions. If we are to conduct inference with these sentences then we may have to add new sentences to the set during inference, so we had better consider the set as containing all sentences constructable from the n propositions. If all sentences are put in conjunctive normal form we can see that there are $2^{2^n}$ different sentences altogether whose probability needs to be stored. However, I am reliably informed* that the probabilities of all these sentences can be recovered provided the probabilities of all the $2^n$ clauses is known. To store a decimal number of m digits requires 10.m bits, thus $10.m.2^n$ are required altogether. To record a probability of m digits in an incidence requires a sample space of size $10^m$, i.e. $10^m$ bits. But the incidences (and hence probabilities) of all sentences of the set can be recovered from the incidences of the n propositions, so only $n \times 10^m$ bits are required altogether. In a typical expert system m will be very small compared with n (uncertainty measures are usually subjective so a value of m>2 would be spurious accuracy, whereas n will have the same order of magnitude as the number of production rules), so $10^m.n \ll 10.m.2^n$. Therefore, the minimum storage required for the incidences is significantly less than that required for probabilities.

---

*D.M. Titterington, Personal Communication.

## 4. Inference Under Uncertainty

Any calculus for uncertainty reasoning needs to provide some mechanism for inheriting uncertainty values from the hypothesis of an inference step to the conclusion, i.e. if we know the uncertainty of A and A->B then we need to be able to calculate the uncertainty of B when modus ponens is applied. By analogy with the truth functionality of Propositional Logic, we will call this property of a calculus, **proof functionality**.

However, this proof functional criterion is too strong and must be relaxed in general. Consider the case of modus ponens, B may be derivable in several ways, e.g. from C and C->B. Any particular derivation can only provide a lower bound on the certainty of B. But we would like to make this lower bound as tight as possible, i.e. to make the calculus as proof functional as possible.

In Incidence Calculus the uncertainty of a sentence is its incidence. In general, if A |- B is a rule of inference of a logical system then all we can infer is that $i(A) \subseteq i(B)$; if A is true in some point then B will be true in that point. Each derivation of B provides a new lower bound for its incidence. The greatest lower bound is found by taking the union of these lower bounds. This is legitimised by the set theoretic theorem,

$$L_1 \subseteq I \ \& \ L_2 \subseteq I \rightarrow L_1 \cup L_2 \subseteq I$$

where I is the incidence and $L_1$ and $L_2$ two lower bounds.

The maintenance of a lower bound of the true incidence is in the same spirit that MYCIN amalgamates the certainty factors calculated from different derivations of the same conclusion, except that the MYCIN amalgamation algorithm is ad-hoc whereas ours is justified by set theory.

In some cases inference steps are proof functional, e.g. when deriving A&B from A and B then $i(A\&B)=i(A)\cap i(B)$. Note that in a probability calculus

$$p(A \ \& \ B) = p(A).p(B) + c(A,B).\sqrt{p(A).p(\tilde{A}).p(B).p(\tilde{B})}$$

where c(A,B), the correlation between A and B, can vary between -1 and +1, so for given p(A) and p(B), p(A&B) can take a range of values. Thus a probability calculus would not be proof functional in this case. We can rephrase this observation by saying that in order for an uncertainty mechanism to be proof functional for this rule of inference it is necessary that & be truth functional with respect to the uncertainty measure. & is truth functional with respect to incidences but not probabilities. Similar remarks will hold for any rule of inference in which the hypotheses are all subformulae of the conclusion. This emphasises the importance of the connectives being truth functional with respect to the uncertainty measure.

In general, the lower bounds provided by Incidence Calculus are much tighter than those provided by a purely numeric probability calculus because the connectives are truth functional with respect to incidences but not to probabilities.

The Rules of Inference of Predicate Calculus preserve truth, i.e. if the hypothesis is true in some interpretation then the conclusion is true in that interpretation. Thus we might have a rule A & B |- A, but not a rule A |- A & B. However, in an uncertainty calculus, we can extend the notion of

181

rule of inference to include both of these rules: while the first, given i(A&B), provides a new lower bound for i(A), the second, given i(A) provides a new upper bound for i(A&B). If we are only given upper and lower bounds for the hypotheses then both rules provide both new upper and new lower bounds for their conclusions.

An inference engine for an Incidence Logic can and should exploit this extended notion of rule of inference operating on upper and lower bounds of the incidences of the sentences of the logic. The Legal Assignment Finder, described in [Bundy 84, Bundy 85], provides such an inference engine for Propositional Incidence Logic. It is a extension of the Beth's Semantic Tableau and is shown to be complete for Propositional Incidence Logic in [Bundy 85].

The Legal Assignment Finder incorporates rules of inference corresponding to each of the connectives, for inheriting both upper and lower bounds and going in both directions. That these connectives are truth functional with respect to incidences is a direct contribution to the proof functionality of the inference engine.

The Legal Assignment Finder is input an initial assignment of upper and lower bounds for the incidences of a set of sentences in Propositional Logic. It uses rules of inference to specialize this initial assignment to just those legal assignments of incidences to these sentences. Suppose F is an assignment, then it defines the functions $\sup_F$ and $\inf_F$ from sentences to incidences. The former gives the upper bound and the latter the lower bound.

For instance, there is a rule of inference

And1:     $\sup_G(A) = [\sup_F(A\&B) \cup w\backslash\inf_F(B)] \cap \sup_F(A)$

where assignment G is a specialization of assignment F. The justification of this rule is that since

$i(A\&B) \subseteq \sup_F(A\&B)$ and
$w\backslash i(B) \subseteq w\backslash\inf_F(B)$     then
$i(A\&B) \cup w\backslash i(B) \subseteq \sup_F(A\&B) \cup w\backslash\inf_F(B)$

But $i(A) \subseteq i(A\&B) \cup w\backslash i(B)$
and $i(A) \subseteq \sup_F(A)$

Therefore: $\sup_G(A) = [\sup_F(A\&B) \cup w\backslash\inf_F(B)] \cap \sup_F(A)$

The Legal Assignment Finder has 4 such rules for negation and 6 each for conjunction, disjunction and implication.

## 5. Assigning Incidences

To initialize an incidence based system, sentences must be given as axioms and incidences must be assigned to them. [Bundy 84, Corlett & Todd 85] discusses how incidences may be calculated from user assigned probabilities and correlations, if available.

An alternative solution is to assign incidences directly. These incidences might

182

be subjective or objective. Subjective incidences would be estimated by the expert or user, just as uncertainty numbers usually are in expert systems. Objective incidences would be provided by data from field studies, just as objective probabilities are provided in Bayesian decision aids.* The human provider of subjective incidences would have to be presented with some sample space, w, of incidents from which to chose. In practice, w would have to be small and might be presented in some graphic form, e.g. a screen divided up into an array. It would help if the incidents were meaningful to the human and related to the domain, e.g. days of the week. The Inconsistency Detection Algorithm could track the assignment and warn of any inconsistency. Empirical studies are required to see whether this is feasible. It is known that humans often object to providing uncertainty numbers; would incidences be better or worse?

The provision of objective incidences involves less statistical analysis than the provision of objective probabilities. Each set of identical experiments becomes an incident and is stored in the incidence of those formulae that were true in the experiments. One does not need to calculate the probability of the formulae, but one does need to estimate the probability of the individual incidents. If each experiment is considered equi-probable then the probability of each incident can be calculated from the number of experiments in its set. One also needs to assume that these incidents cover all the cases.

## 6. Conclusion and Further Work

In this note we have described a mechanism, Incidence Calculus, for incorporating probabilistic reasoning in an inference system. This mechanism is based on the assignment of sets of points to sentences, rather than the normal technique of assigning numbers to sentences. Our mechanism provides a probabilistic logic with truth functional connectives. A purely numeric mechanism cannot provide this. Because of this property, Incidence Calculus provides tighter bounds on the probabilities of inferred sentences than could be provided by such a numeric mechanism.

Incidence Calculus can be implemented reasonably efficiently using bit strings. It is then a more efficient technique for storing the probabilities of sentences than a purely numeric mechanism, especially if the dependencies between sentences must also be stored.

We have designed and implemented a complete inference engine for Propositional Incidence Logic, the Legal Assignment Finder, and outlined the extensions required to extend this into an inference engine for Predicate Incidence Logic. More work is required to explore the theoretical properties of this extended engine, e.g. completeness and soundness, and to implement and test it.

Incidence Calculus can be readily adapted to Shafer-Dempster Theory, i.e. to returning an interval of probabilities rather than a single value.

Some of the mechanisms described in this paper have been implemented. Further implementation and testing is required.

---

*I am indebted to Richard O'Keefe for suggesting using objective incidences.

**References**

[Bundy 84]      Bundy, A.
                Incidence Calculus: A Mechanism for Probabilistic Reasoning.
                In Aiso, H. (editor), *Proceedings of the Fifth Generation Computer Systems Conference*, pages 166-174. November, 1984.
                To appear in the Journal of Automated Reasoning.

[Bundy 85]      Bundy, A.
                *Correctness criteria of some algorithms for uncertain reasoning using Incidence Calculus*.
                Research Paper, Dept. of Artificial Intelligence, Edinburgh, 1985.
                Submitted to the Journal of Automated Reasoning.

[Corlett & Todd 85]
                Corlett, R.A. and Todd, S.J.
                A Monte-Carlo approach to uncertain inference.
                In Ross, P. (editor), *Proceedings of AISB-85*, pages 28-34. 1985.

[Duda et at 78] Duda, R.O., Hart, P.E., Nilsson, N.J. and Sutherland, G.L.
                Semantic network representations in rule-based inference systems.
                In Waterman, D and Hayes-Roth, F. (editor), *Pattern- directed inference systems*, pages 203-221. Academic Press, 1978.

[Feller 68]     Feller, W.
                *An introduction to Probablity Theory and its applications*.
                John Wiley & Sons, 1968.
                Third Edition.

[Lowrance & Garvey 82]
                Lowrance, J.D. and Garvey, T.D.
                Evidential reasoning: A developing concept.
                In *Proceedings of the International Conference on Cybernetics and Society*, pages 6-9. IEEE, 1982.

[Shortliffe 76] Shortliffe, E.H.
                *Computer-based medical consultations: MYCIN*.
                North Holland, 1976.

[Zadeh 81]      Zadeh, L.
                PRUF - a meaning representational language for natural languages.
                In Mamdani, E. and Gaines, B. (editors), *Fuzzy reasoning and its applications*, . Academic Press, 1981.